\documentclass[letterpaper, 10 pt, conference, keeplastbox]{ieeeconf}  

\IEEEoverridecommandlockouts                              % This command is only needed if 
\usepackage{graphics} % for pdf, bitmapped graphics files
\usepackage{graphicx}
\usepackage{pict2e} %to put text into figure
\usepackage{subfig}
\usepackage{amsmath} % assumes amsmath package installed
\usepackage{pict2e} %to put text into figure
\usepackage{xcolor} % for text color
\usepackage{booktabs}
\usepackage{graphicx}
\usepackage{textcomp} % now we can use /texttrademark
\usepackage{siunitx} % space between numbers and units
\usepackage{flushend}
\usepackage{hyperref}

\title{\LARGE \bf A Low-Cost, Easy-to-Manufacture, Flexible, Multi-Taxel Tactile Sensor \\ and its Application to In-Hand Object Recognition}
\author{Tessa J. Pannen* \qquad Steffen Puhlmann* \qquad Oliver Brock% <-this % stops a space
\thanks{
*These authors contributed equally to this work. \newline
 All authors are with the Robotics and Biology Laboratory, Technische Universit\"{a}t Berlin, Germany. We gratefully acknowledge financial support by the European Commission (SOMA, H2020-ICT-645599), the Deutsche Forschungsgemeinschaft (DFG, German Research Foundation) under Germany's Excellence Strategy - EXC 2002/1 "Science of Intelligence" - project number 390523135 and German Priority Program DFG-SPP 2100 ``Soft Material Robotic Systems'' - project number 405033880.}%
}

\begin{document}
\maketitle
\thispagestyle{empty}
\pagestyle{empty}

%======================================================================
\begin{abstract}
%======================================================================
Soft robotics is an emerging field that yields promising results for tasks that require safe and robust interactions with the environment or with humans, such as grasping, manipulation, and human-robot interaction. Soft robots rely on intrinsically compliant components and are difficult to equip with traditional, rigid sensors which would interfere with their compliance.
We propose a highly flexible tactile sensor that is low-cost and easy to manufacture while measuring contact pressures independently from 14 taxels. The sensor is built from piezoresistive fabric for highly sensitive, continuous responses and from a custom-designed flexible printed circuit board which provides a high taxel density. From these taxels, location and intensity of contact with the sensor can be inferred. In this paper, we explain the design and manufacturing of the proposed sensor, characterize its input-output relation, evaluate its effects on compliance when equipped to the silicone-based pneumatic actuators of the soft robotic RBO~Hand~2, and demonstrate that the sensor provides rich and useful feedback for learning-based in-hand object recognition. 
\end{abstract}

%======================================================================
\section{Introduction}\label{ch:intro}
%======================================================================

Soft robotic hands demonstrate robust grasping and manipulation capabilities thanks to their intrinsic compliance~\cite{sieverling2017_cerrt, gupta2016learning, pall2018_concerrt}. However, sensorization of these soft hands is challenging, as suitable sensors must provide sensing abilities without detrimental effects on compliance. Soft sensors therefore need to be based on flexible materials, however, useful designs that provide rich feedback often rely on complex fabrication or costly components.

\begin{figure}[!t]
  \centering
  \includegraphics[width=0.8\linewidth]{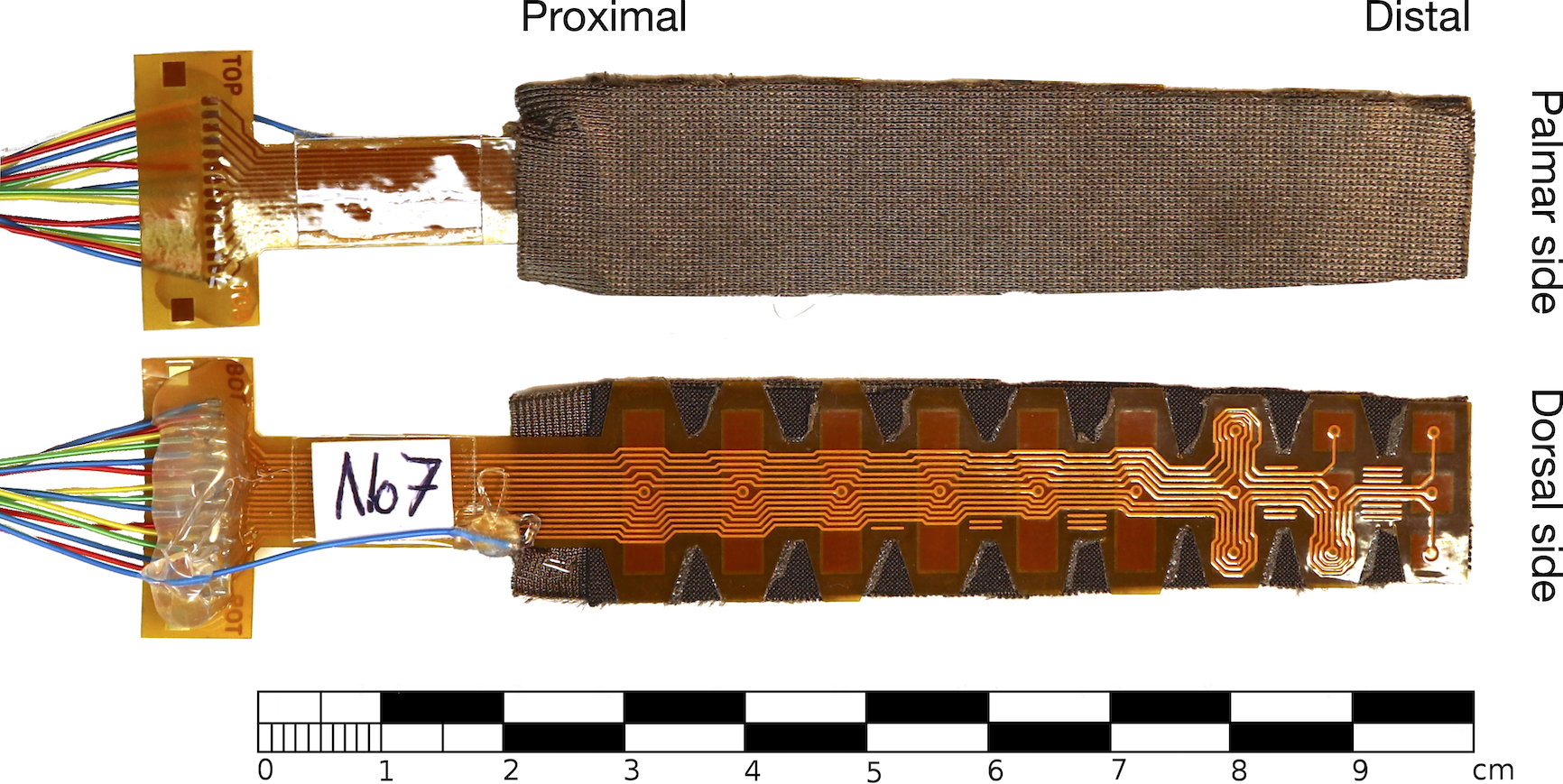}
  \caption{A flexible tactile sensor:
    palmar side shows the grounding electrode made of Medtex~P130
    fabric (top), dorsal side shows the flexPCB integrating
    electrodes and wiring (bottom). The sensor has 14 independent
    taxels (the most proximal electrode/taxel is not used).}
  \label{fig:sensor}
\end{figure}

We propose a novel, highly compliant tactile sensor (Fig.~\ref{fig:sensor}) suitable for soft robotics that is low-cost and very easy to manufacture while providing rich sensory feedback with high spatial resolution, enabling learning-based object recognition. The sensor relies on piezoresistive fabric which changes its electrical resistance under contact pressure. It is highly sensitive while measuring a wide range of contact pressures. Its sensitivity and measuring range can be adjusted to support a large variety of applications with different contact intensities. To effectively distinguish contact patterns, our sensor possesses 14 taxels realized by a custom-designed flexible printed circuit board (flexPCB). This flexPCB tightly integrates electrodes and wiring in a thin layout.  

We evaluate the sensor in grasping experiments with the highly compliant, anthropomorphic RBO~Hand~2~\cite{deimel_novel_2016} and show that it provides useful tactile information, enabling classification of grasped objects. We also show that the sensor has only limited effect on the hand's compliance, so that passive shape adaptation during grasping is still attainable. Finally, we compare object classification based on the tactile sensor to classification based on air pressure sensing in the hand's pneumatic actuators. Our results show that thanks to its superior spatial resolution, our sensor outperforms pressure-based object classification, highlighting that our cost-effective and simple design yields rich feedback.

%======================================================================
\section{Related Work}
\label{ch:soa}
%======================================================================

Related works presented a wide range of tactile sensors based on a variety of technologies~\cite{Yousef2011, lu2014flexible, Kappassov2015}. We now discuss promising research that contributes to achieving dense tactile sensing while limiting detrimental effects on the compliance of a sensorized soft actuator.

Microchannels filled with liquid metal change their resistance under deformation, enabling the measurement of multi-axis strain and pressure~\cite{Park2012, Fang2019}. Several such sensors can be combined to infer contact information on the entire actuator, including deformation, forces, and contact location~\cite{WallBrock2017,WallBrock2019}. These sensors have only minor effect on the compliance of a sensorized actuator but they have not yet been shown to provide dense tactile sensing.

Acoustic sensing relies on sound propagating through a soft continuum actuator to infer information about contact location, contact intensity, and contacted material~\cite{zoller2018acoustic,zoller2020acoustic}. Since the required microphone and optional speaker can be embedded into the cavity of the pneumatic actuator, the effect on compliance is negligible. However, only single contact locations can be measured. The sensor presented in this paper allows concurrent sensing of many contacts.

Optical tactile sensors use cameras to record deformation-based changes of marker patterns~\cite{ward2018tactip,ward2020miniaturised} or of the surface relief~\cite{yuan2017gelsight} on the inside of soft actuators. Embedding these cameras requires a rigid base for the soft actuator, limiting the compliance and design choices.

Elastomers enriched with carbon black particles change their resistance under deformation~\cite{Lacasse2010, stassi2014flexible}. However, these particles can affect the elastomer's mechanical properties, so that increasing sensitivity comes at the expense of lower flexibility~\cite{stassi2014flexible}.

A different design for soft tactile sensors sandwiches piezoresistive foil or fabric between electrode layers. Individual patches of the piezoresistive material encode the taxel structure~\cite{Low2016,Butt2019}. Placed on a soft robotic finger, these sensors can infer gasping forces and locations. However, they have relatively low tactile density and published designs often have external cable routing, possibly complicating operation in proximity to obstacles.

Alternatively, the electrode layers encode the taxel structure~\cite{Buscher2015,Chen2018}. The fabrication of these sensors is often complex, for example requiring etching of the electrode layers. However, these designs offer relatively high taxel density while promising not to reduce compliance substantially. We therefore base our work on this kind of design. Our sensor, described in the next section,
offers simple manufacturing and dense tactile sensing, overcoming the problems of existing tactile sensors based on this design.

%======================================================================
\section{Tactile Sensor Design and Manufacturing}\label{ch:design}
%======================================================================

\begin{figure}[!t]
  \centering
  \includegraphics[width=0.7\linewidth]{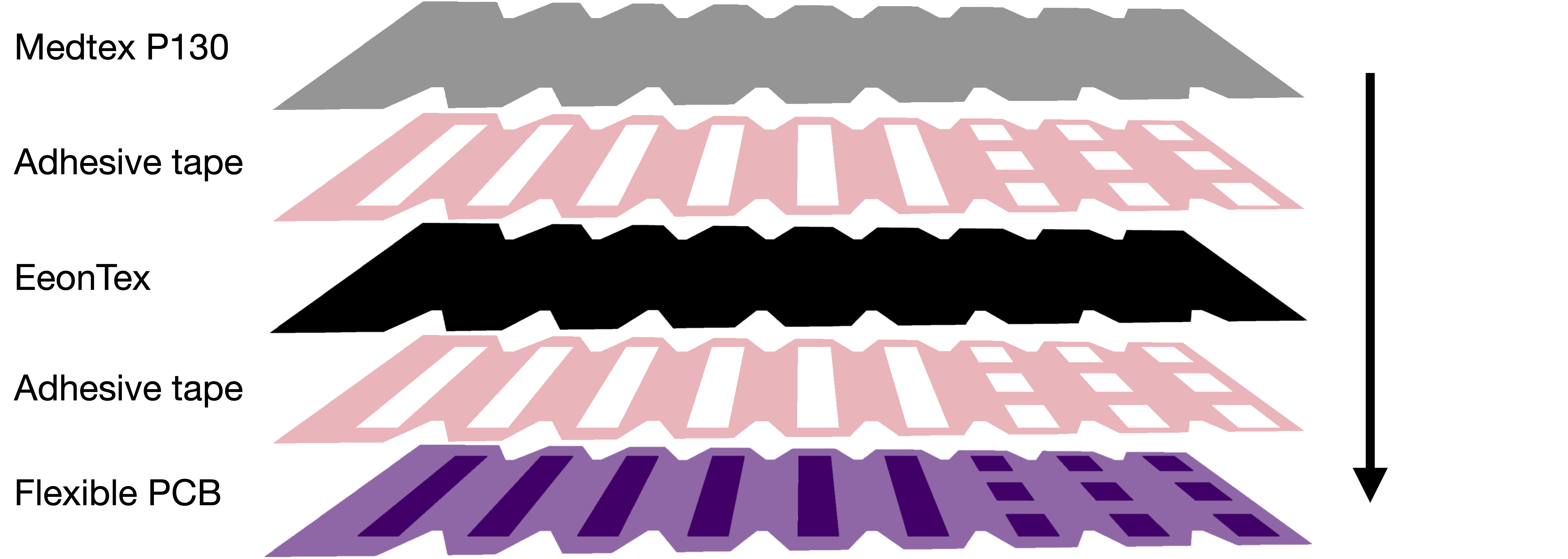}
  \caption{Fabrication of the tactile sensor: a custom-designed flexPCB (bottom) which
    integrates fourteen independent taxels and wiring, a laser-cut sheet of EeonTex
    piezoresistive fabric (center) which changes its electrical resistance
    under contact pressure, and a laser-cut sheet of silver-coated nylon fabric (top) which serves
    as a grounding electrode are connected by laser-cut double-sided adhesive tape.
  }
  \label{fig:sensorlayers}
\end{figure}

Our tactile sensor should have only little effect on the robot's morphology and compliance. It should exhibit a high spatial resolution and provide continuous responses for a large range of contact pressures with high sensitivity. At the same time, it should be based on low-cost materials and very simple to manufacture. We now describe how we achieve these design goals.

The sensor is based on a piezoresistive material which is sandwiched between two electrode layers for reading-out its electrical resistance (Fig.~\ref{fig:sensor}). We divide the sensor into different regions (taxels) for which we determine the resistance independently (electrodes of taxels are connected in parallel). To minimize the sensor's effect on morphology and compliance, we choose thin and flexible materials, including conductive fabrics and a flexPCB. 

The sensor design is based on EeonTex, a commercially available piezoresistive fabric which is only \SI{0.4}{mm} thick and made of nylon fabric coated with conductive polymers. EeonTex enables highly sensitive responses, changing its electrical resistance by more than 99\% under contact pressure. This resistance is measured at different regions via multiple electrodes (taxels) on a flexPCB and a single grounding electrode based on Medtex~P130, conductive fabric made of silver-plated nylon (Fig.~\ref{fig:sensorlayers}). 

The flexPCB provides a high spatial resolution of up to \SI{4.5}{mm} at the fingertip, realized by 14~taxels while integrating electrodes and cabling in a custom-designed, easily accessible and low-cost component. At the fingertip, it has nine square electrodes (\SI{3}{mm} side length) and five rectangular taxels \mbox{(3 $\times$ \SI{13}{mm})} located proximally (Fig.~\ref{fig:sensorlayers}). The flexPCB enables a clear and clutter-free design by keeping the sensing surface void of solder joints and shifting cabling connections out of the actuator's workspace. 
These design choices result in a thin, compliant, and highly sensitive sensor whose thickness ranges between \SI{1.0}{mm} and \SI{1.2}{mm} due to small variations in manufacturing.

Manufacturing the sensor is very simple and fast, allowing for rapid prototyping and easy replacements. The fabrics and the flexPCB are connected with double-sided adhesive tape (Fig.~\ref{fig:sensorlayers}). The fabrics and the tape are laser-cut with cut-outs in the tape that correspond to the shapes and locations of the electrodes on the flexPCB. Finally, we solder cables to wire outlets at the bottom of the flexPCB.  Following this procedure, a sensor can be assembled within minutes. The cost of the sensor is dominated by the commercially manufactured flexPCB while the cost of a single piece drops substantially when ordering large quantities. In our case, a batch size of 50 pieces resulted in a sensor costing less than US\$~5, demonstrating the cost-effectiveness of our design.

The sensor is glued to the palmar side of the silicone-based finger of the RBO~Hand~2 using \mbox{Sil-Poxy} silicone adhesive, with the \mbox{flexPCB} facing the actuator.
Since this wet adhesive would soak into the fabrics, we cover the sensor with a thin, water resistant sleeve made of \mbox{Dragon~Skin~10} silicone.

%======================================================================
\section{Sensor Characterization}\label{sec:characterization}
%======================================================================

\begin{figure}[!t]
  \centering
  \includegraphics[width=\linewidth]{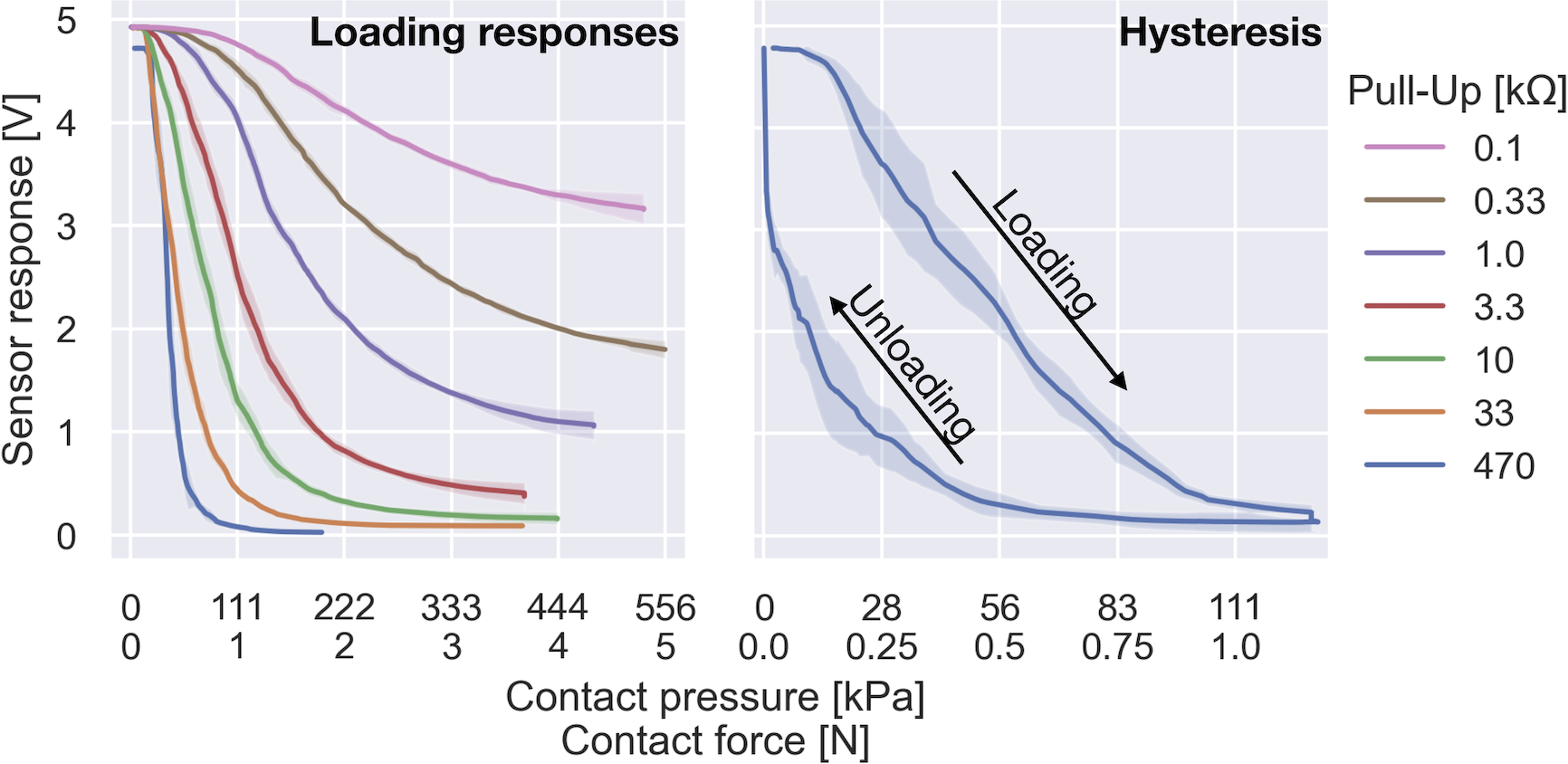}
  \caption{The sensor response can be adjusted by the
    choice of pull-up resistance, trading measuring range with
    sensitivity~(left). Sensor shows repeatable hysteresis due to relaxation time of the piezoresistive fabric~(right).  Solid lines show mean responses over ten trials, intervals indicate
    standard deviation. %A square probe-tip with \SI{3}{mm} side length is used for all cases.
    }
  \label{fig:resistors}
\end{figure}

We now characterize the sensor's response behavior and explain how it can be adjusted to match task requirements, before we analyze the sensor's effect on compliance.

%----------------------------------------------------------------------
\subsection{Response Behavior}
%----------------------------------------------------------------------

We probe the sensor's response behavior to analyze its sensitivity, measuring range, and hysteresis. For this, we apply normal forces to a fingertip taxel, using a square probe-tip of \SI{3}{mm} side length which is attached to a force-torque sensor. Contact forces and sensor responses are recorded at \SI{20}{Hz}, with each recording lasting \SI{3.5}{s}.

In response to contact, the electrical resistance of the EeonTex fabric changes at the location of the taxel. Instead of measuring this resistance directly, we use a voltage divider and infer the response indirectly by measuring the voltage between the conductive Medtex~P30 fabric and the electrode on the flexPCB that corresponds to this taxel.
The relationship between contact pressure and output voltage is described by the characteristic curve. The sensor is most sensitive at the center of its measuring range where the slope of the characteristic curve is steepest. 

As with other piezoresistive tactile sensors, the choice of pull-up resistance modulates its response, with a large resistances resulting in a narrow measuring range and large sensitivity, and vice versa. The ability to adjust its behavior to capture expected contact intensities for different tasks makes our design highly versatile.

Similar to other piezoresistive sensors~\cite{Yousef2011, Kappassov2015,  Buscher2015},
our sensor exhibits hysteresis, because the EeonTex fabric needs time to relax after contact-based compression. The larger the hysteresis, the smaller a sensor's sensibility for high frequency changes. We measured an average relaxation time of ca.~\SI{0.41}{s} after unloading has ended. 

Figure~\ref{fig:resistors} shows measured responses during the loading process for different pull-up resistors, and the hysteresis during the loading/unloading cycle. 
These responses are averaged across ten trials each and highly predictable with a standard deviation of only \SI{0.29}{V} on average across contact pressures and \SI{0.67}{V} at maximum. We did not experience cross-talk at neighboring taxels.

%----------------------------------------------------------------------
\subsection{Effect on Compliance}
\label{subsec:compliance}
%----------------------------------------------------------------------

\begin{figure}[!t]
  \centering
  \includegraphics[width=\linewidth]{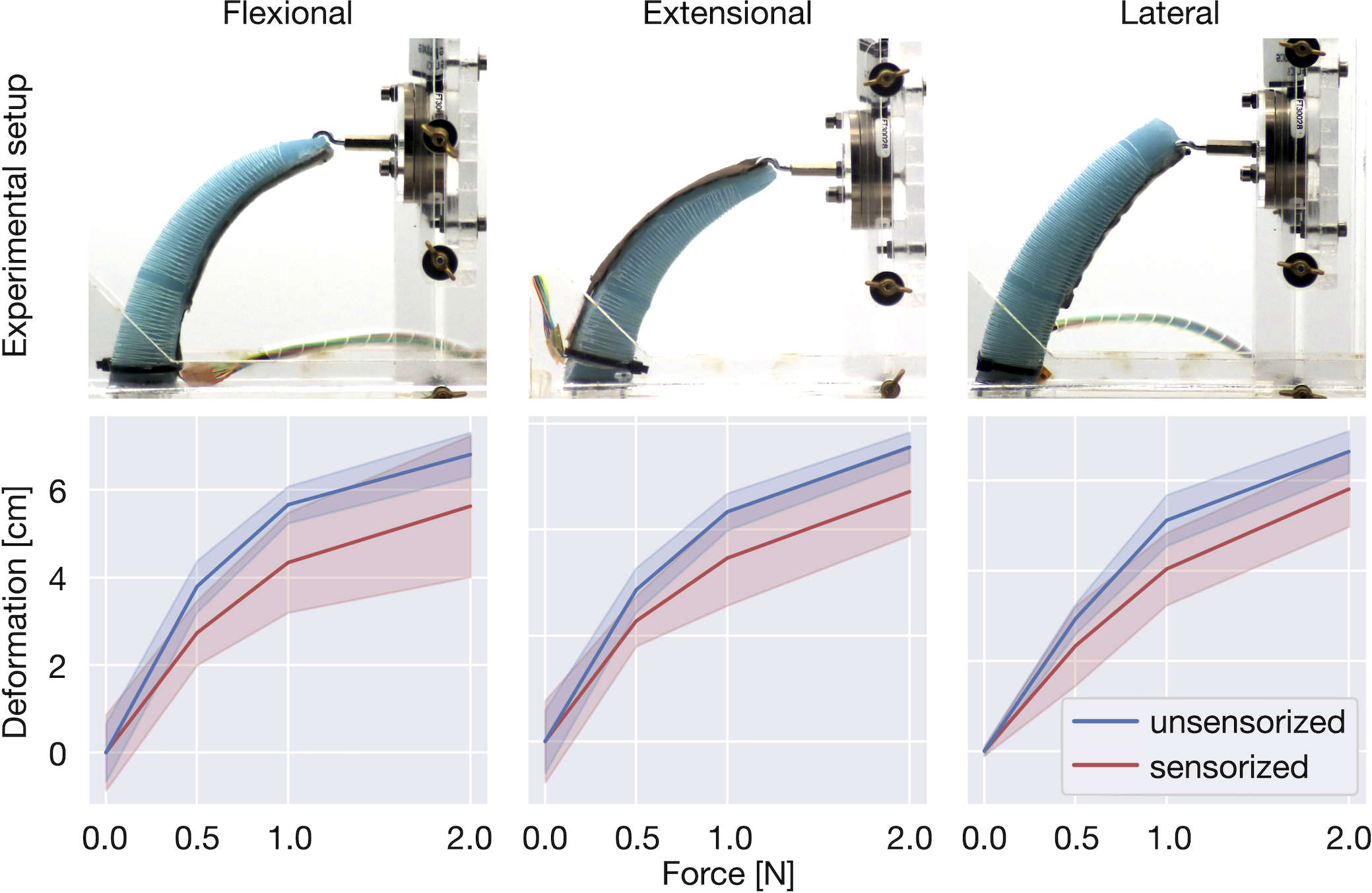}
  \caption{The tactile sensor has only limited effect on finger compliance.
    Experimental setup for measuring finger deformations due
    to horizontal pulling forces in flexional (top left), extensional
    (top center), and lateral (top right) direction with a force-torque sensor 
    which is hooked into the fingertip.  Bottom
    row: fingertip displacements along horizontal axis for different
    pulling forces. Solid lines show mean deformation over 15 trials (five fingers, three trials per finger), intervals  indicate standard deviations.}
  \label{fig:compliance}
\end{figure}

A sensor that substantially reduces compliance would impair the grasping and manipulation capabilities of a soft hand. Therefore, we analyze our sensor's effect on compliance when attached to the soft fingers of the RBO~Hand~2.

We compare five sensorized to five non-sensorized fingers when applying pulling forces of \SI{0.5}{N}, \SI{1}{N}, and \SI{2}{N} in flexional, extensional, and lateral directions (Fig.~\ref{fig:compliance}). To test a finger, it is fixated at its base and a force-torque sensor is connected to its fingertip via a metal hook. The force-torque sensor is pulled away from the finger until a specific pulling force is reached while its vertical position is adjusted to maintain horizontal pulling forces. We determine the resulting deformations by measuring the distance along the horizontal axis between the fingertip position in its relaxed and deformed pose. For each finger, we repeat this process three times for each pulling force in each direction.

Figure~\ref{fig:compliance} shows the resulting finger deformations for the different forces and directions. The average distance between fingertip positions is \SI{11.9}{mm} for flexional, \SI{7.7}{mm} for extensional, and \SI{8.4}{mm} for lateral forces. On average, the sensorized finger deforms about 19.5\% less than its non-sensorized counterpart. The measurable effect on compliance is not surprising since we added material to the finger. However, this difference does not impair the hand's ability to passively adapt its morphology to the shape of grasped objects, as we show in the following section. The limited effect on compliance---despite the non-extensibility of the sensor---stems from the fact that the finger's palmar side is reinforced by fabric that itself is non-extensible so that it bends upon inflation. The ability to quantify the loss of compliance due to sensorization allows making informed adjustments to the finger's geometry or to the choice of materials in order to compensate for this loss, if needed.

%======================================================================
\section{In-Hand Object Recognition} \label{ch:recognition}
%======================================================================

We demonstrate our sensor's ability to provide useful contact information by enabling a sensorized hand to reliably recognize grasped objects. High classification rates indicate a sensor's ability to capture distinctive object features from tactile information. We compare object recognition performance based on only tactile sensor information to a baseline case when only air pressure information from the pneumatic actuators is considered. Furthermore, we analyze failure cases with respect to physical similarities between objects and investigate how different levels of spatial resolution contribute to successful object recognition.

\begin{figure}[!t]
  \centering
  \includegraphics[width=\linewidth]{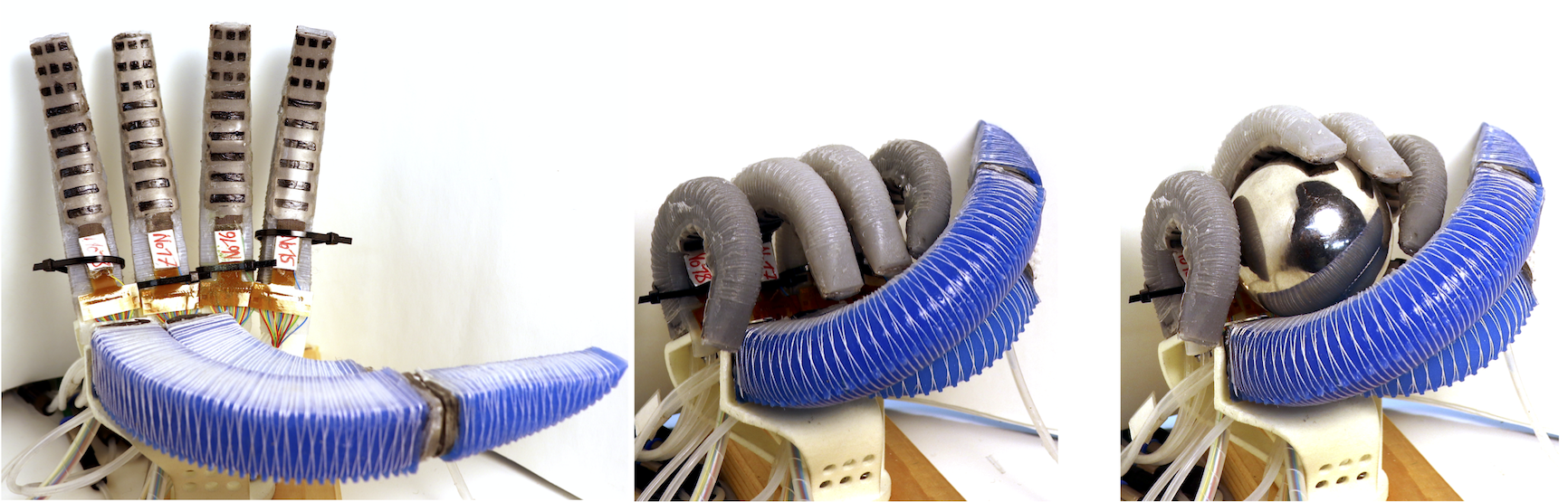}
  \caption{Compliant, sensorized RBO~Hand~2 passively adapts its shape
    to the grasped object. Left: non-inflated hand. Center: closed
    hand in absence of an object after performing the inflation
    pattern used for all grasps. Right: hand-object configuration
    after compliantly grasping a sphere (not part of the object set).}
  \label{fig:grasping}
\end{figure}

%----------------------------------------------------------------------
\subsection{Experimental Setup}
%----------------------------------------------------------------------

We equip the soft pneumatic RBO~Hand~2 with tactile sensing by attaching our sensor to the object-facing (palmar) side of its four silicone-based fingers (Fig.~\ref{fig:grasping}).
For comparing our sensor to a baseline case, we also record the pressure inside each air chamber with Freescale MPX4250 sensors.  Each finger has two of these air chambers: one at its base (proximal) and one at its tip (distal). The actuators for thumb movement do not provide any data, but are used for grasping. 

We adjust the sensor's measuring range and sensitivity to match expected contact intensities. We found that during grasping, contact pressures remain in more than 97.9\% of the cases below 75~kPA. Based on the sensor's response behavior (Fig.~\ref{fig:resistors}), we choose a pull-up resistance of \SI{470}{k\Omega} so that these pressures are well within the sensor's measuring range.

\begin{figure}[t!]
  \centering
  \includegraphics[width=\linewidth]{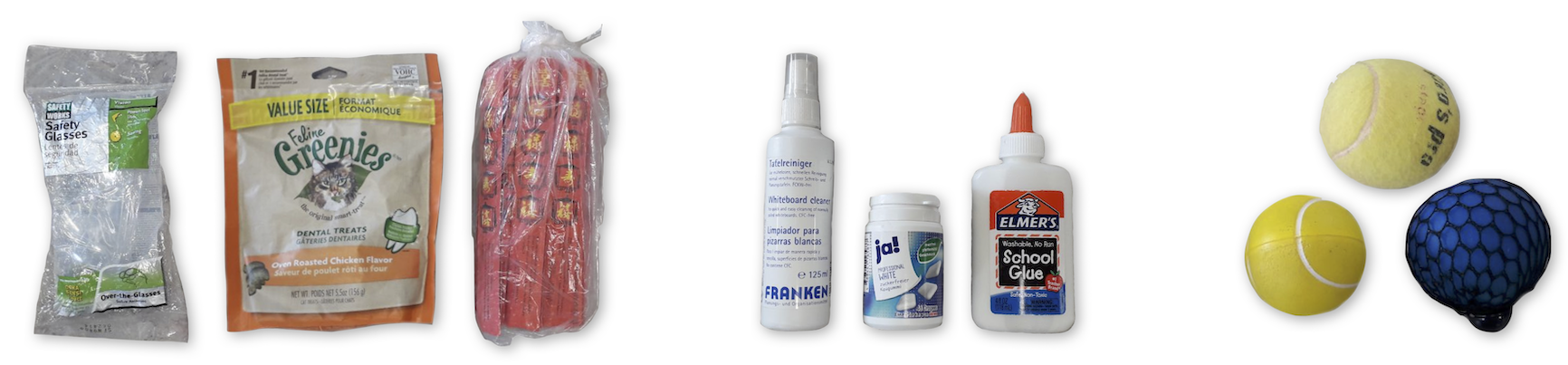}
  \caption{Object set with nine objects from three object
    groups, from left to right: objects in packaging (safety glasses, cat treats, chopsticks),
    plastic bottles (whiteboard cleaner, gum container, wet glue), and
    spherical objects (soft ball, tennis ball, gel ball).}
  \label{fig:objects}
\end{figure}

To evaluate the ability of the sensorized hand to reliably recognize every-day objects,
we chose objects with diverse physical properties that fall into three groups regarding shape and softness (Fig.~\ref{fig:objects}). Objects within the same group exhibit highly similar physical properties and are therefore difficult to distinguish. The three object groups are  i)~objects that can move inside deformable packaging, including cat treats, safety glasses, and chopsticks, ii)~rigid plastic bottles, including a whiteboard cleaner, a gum container, and wet glue, and iii)~spherical objects, including a tennis ball, a soft ball, and a gel ball which all have similar diameters, but different degrees of softness.

%----------------------------------------------------------------------
\subsection{Data Acquisition}
%----------------------------------------------------------------------

To record sensory feedback, one of the objects is placed inside the open hand with its weight resting on the palm which is supported by rigid material, preventing the object from falling out and its weight from influencing contact measurements (Fig.~\ref{fig:grasping}). Objects are placed in random orientation, resulting in a variety of contact patterns for the same object. Hand closure brings the tactile sensors in contact with the object while all of the fingers and also the thumb are actuated (Fig.~\ref{fig:grasping}).  
The same closing-synergy, based on pre-defined air masses, is executed for each object. Because of the hand's intrinsic compliance, it passively adapts to the shape of the object, resulting in distinct contact patterns. Once the hand is fully closed, a delay of three seconds ensures that all sensors have stabilized after inflation-dependent disturbances. Responses of the tactile sensor with 14 taxels and of the two air pressure sensors is measured for each of the four fingers, resulting in a data point of $4 \times (14 + 2) = 64$ dimensions. We repeat this process 30~times for each of the nine objects, resulting in 270~data points in total.

%----------------------------------------------------------------------
\subsection{Classification Results}
%----------------------------------------------------------------------

\begin{figure}[!t]
  \centering
  \includegraphics[width=0.9\linewidth]{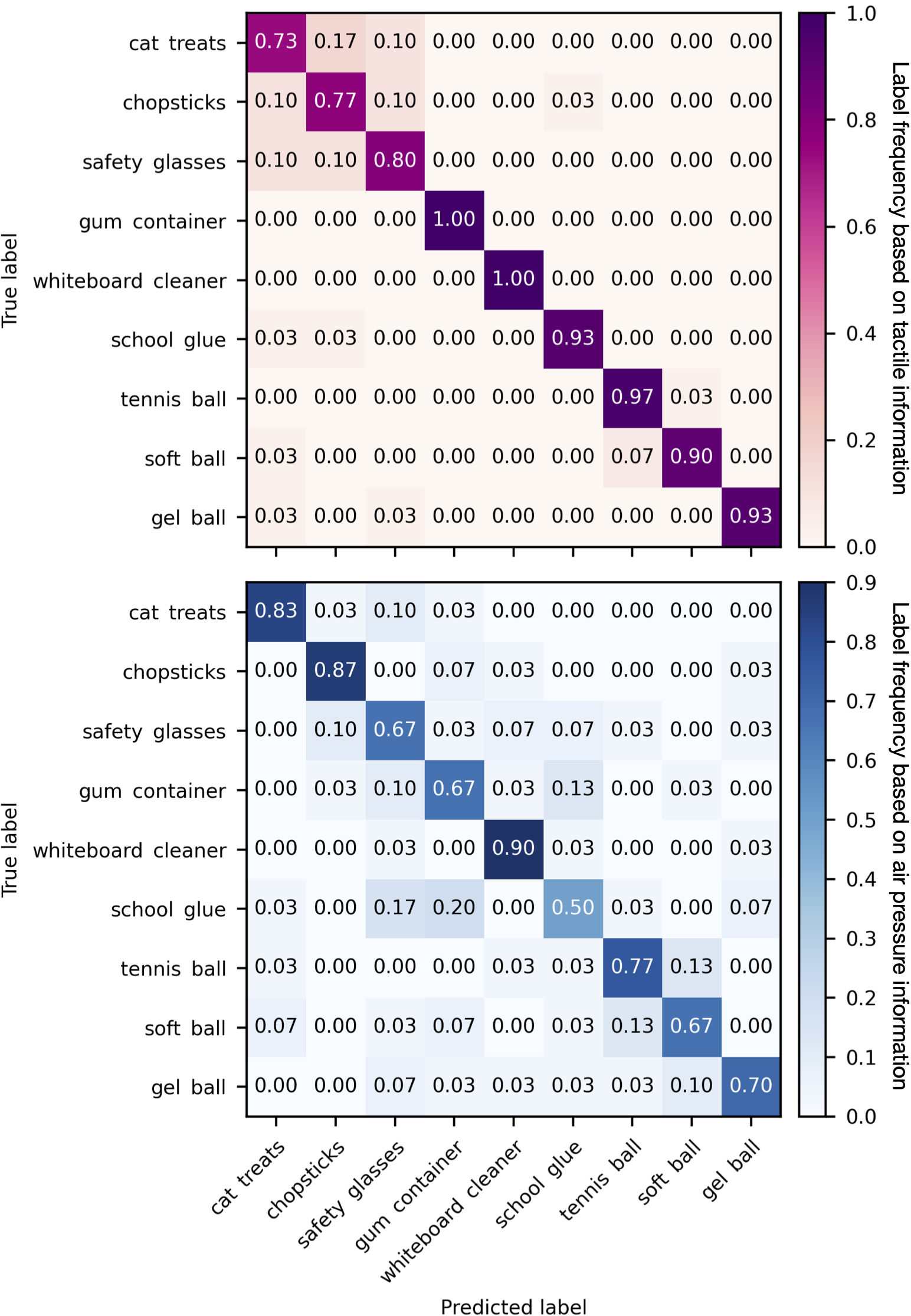}
  \caption{Confusion matrices for object recognition based on
    tactile~(violet) and air pressure~(blue) information: entries
    indicate the frequency of a label being assigned to a specific
    object by the classifier. Results were obtained by leave-one-out
    cross-validation. The tactile sensor exhibits superior performance
    for all objects except cat treats and chopsticks.}
  \label{fig:confusion}
\end{figure}

We analyze recorded contacts for enabling the sensorized hand to correctly classify grasped objects. Specifically, we use machine learning to train a classifier based on the acquired data set. Training was performed based on either only tactile information or only air pressure information (56 and 8 dimensions, respectively). To optimize classification accuracy, we compare different machine learning algorithms and when applicable, perform grid search for hyperparameter optimization. Tested algorithms include \mbox{k-nearest neighbors~(KNN)}, decision trees~(DT), and support vector machines~(SVM) with linear, polynomial and radial basis function (RBF) kernels. Classification performance was evaluated by leave-one-out cross-validation (Table~\ref{tab:my-table}).

\begin{table}[]
\centering
\resizebox{\linewidth}{!}{%
\begin{tabular}{@{}llllll@{}}
\toprule
            & \textbf{KNN} & \textbf{DT} & \textbf{SVM (Linear)} & \textbf{SVM (Polynomial)} & \textbf{SVM (RBF)} \\ \midrule
\textbf{Tactile}      & 77.8         & 73.3        & 88.2                  & 87.4                      & \textbf{89.3              } \\ %\midrule
\textbf{Air Pressure} & 55.2         & 66.3        & 69.3                  & 71.5                      & \textbf{73.0}               \\ \bottomrule
\end{tabular}%
}
\caption{Object recognition performance [\%] of different
  classification algorithms: the SVM with a RBF kernel performs best
  for both sensor modalities.}
\label{tab:my-table}
\end{table}

We found that an SVM with an RBF kernel performs best for both sensory modalities.  In the following, we will refer to this classifier when discussing results~(Fig.\ref{fig:confusion}).
On average, tactile-based classification achieves an accuracy of 89.3\%, which is comparable with performances of other sensorized soft hands~\cite{jiao2020visual, homberg2015haptic}.
In comparison, pressure-based classification achieves only 73.0\%, highlighting the superiority of our tactile sensor. However, for cat treats and chopsticks, air pressure sensing outperforms tactile sensing by 10\%, with 83\% and 87\% compared to 73\% and 77\%, respectively. Both are objects in packaging, indicating that classification performance varies across object groups.

%----------------------------------------------------------------------
\subsection{Failure Case Analysis}
%----------------------------------------------------------------------

Since classification performance varies across object groups, we now investigate failure cases in more detail with respect to physical similarities between objects. For this, we categorize assigned labels into three categories:  i)~label is correct, 
ii)~label is incorrect but part of the same object group as the grasped object,
iii)~label and object group are incorrect.

\begin{figure}[!t]
  \centering
  \includegraphics[width=\linewidth]{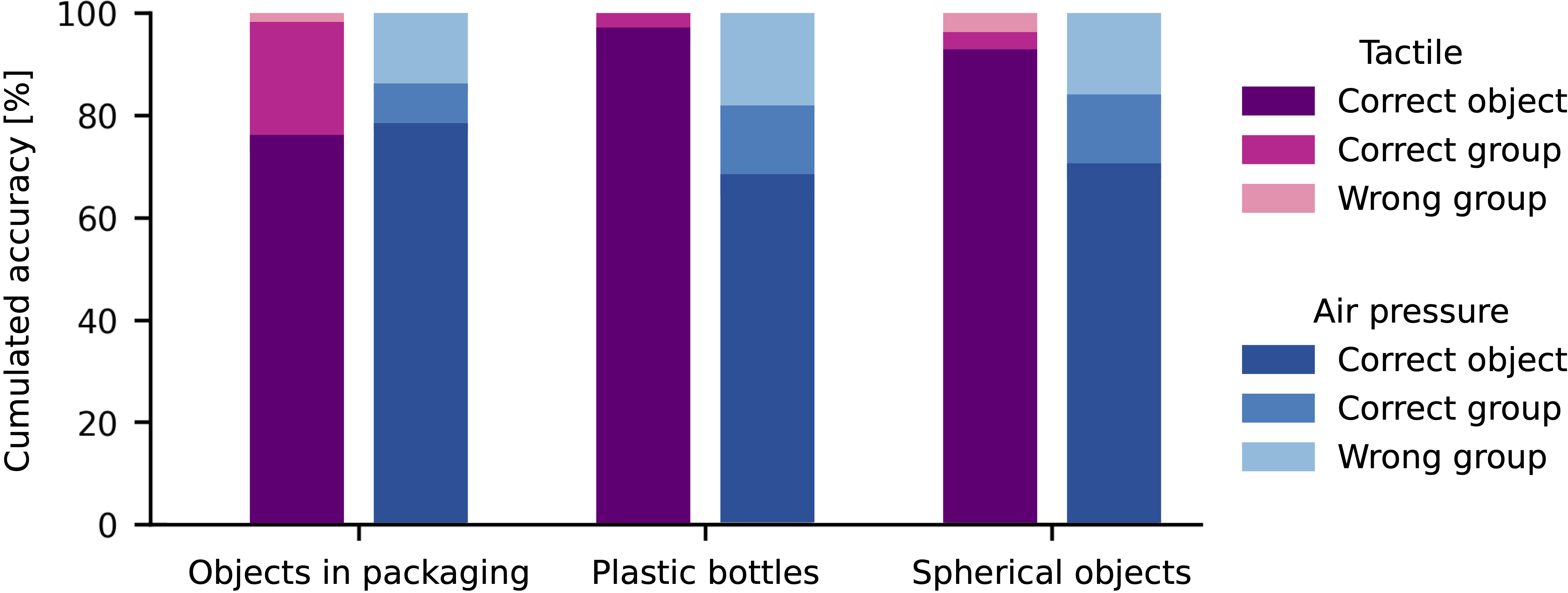}
  \label{fig:analysis-tactile}
  \caption{Classification results for
    tactile~(violet) and air pressure~(blue) sensing:  
    assigned labels are either correct (dark color), incorrect
    but within the same object group as the grasped object (medium
    color), or outside this group (light color).  
    Incorrect tactile-based labels are more frequently within the
    correct object group.}
\label{fig:prediction-analysis}
\end{figure}

The frequencies of these three categories are depicted in Figure \ref{fig:prediction-analysis}. On average, of all incorrectly assigned tactile-based labels, 79.3\% are within the correct object group, compared to only 42.5\% of the incorrectly assigned air pressure-based labels. Thus, not only is tactile sensing superior in correctly assigning labels, but  also in its incorrect labels which are more often part of the correct object group. In particular, for objects in packages, for which air pressure-based classification yields slightly better results, the tactile-based classifier assigns nearly all of the incorrect labels to the correct object group with 95.3\% of the cases, compared to only 36.8\% for air pressure sensing. 

We argue that the tactile sensor's performance for objects in packages is inferior, because these objects have moving parts inside, resulting in a large variety of tactile responses when grasping the same object multiple times. On the other side, the coarse sensing of air pressure sensors is less affected by these movements, as long as object volume and deformation of the compliant fingers remain similar.

%----------------------------------------------------------------------
\subsection{Spatial Resolution}
%----------------------------------------------------------------------

\begin{figure}[!t]
  \centering
  \includegraphics[width=0.8\linewidth]{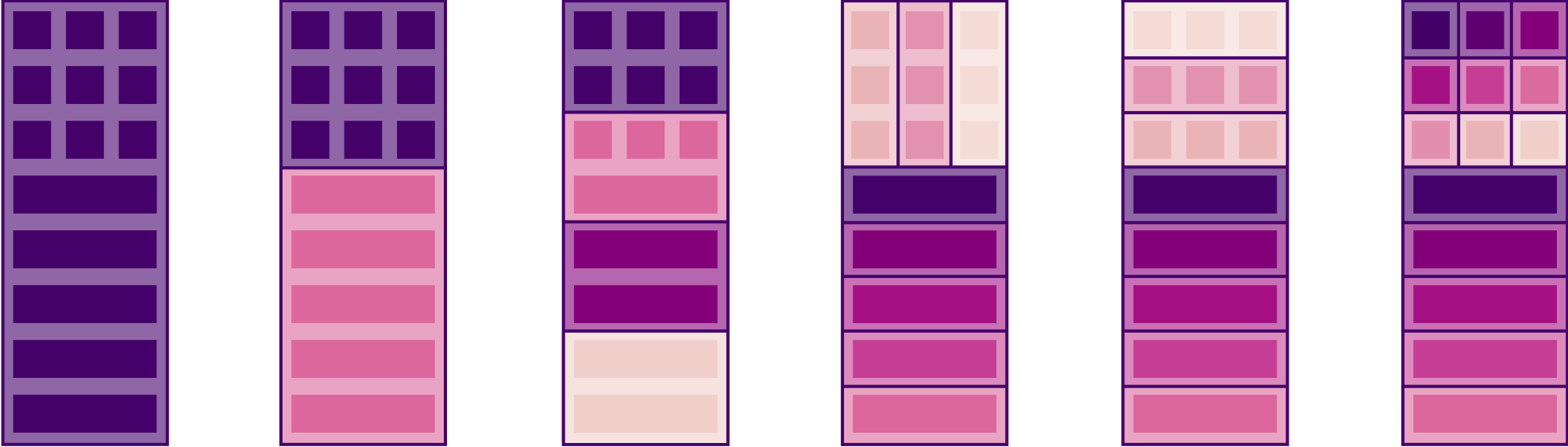}
  \caption{Tactile sensor with different spatial resolutions,
    simulated by averaging over contact information from 
    taxels of the same color, number of taxels from left to right:
    one, two, four, eight with vertically combined fingertip taxels,
    eight with horizontally combined fingertip taxels, fourteen.}
\label{fig:res_tactile}
\end{figure}

We demonstrate that object classification performance improves with increasing spatial resolution. For this, we simulate reduced spatial resolutions of tactile and air pressure sensors by averaging over feedback from multiple taxels, and from the two air chambers, respectively. The reduced tactile resolutions result in taxel configurations as depicted in Figure~\ref{fig:res_tactile}, ranging from a single taxel to fourteen taxels per finger. For the air pressure sensor, this leads to configurations with a single or with two air chambers. We trained classifiers based on the different simulated taxel combination, after optimizing hyperparameters separately for each case.

\begin{figure}[!t]
  \centering
  \includegraphics[width=0.8\linewidth]{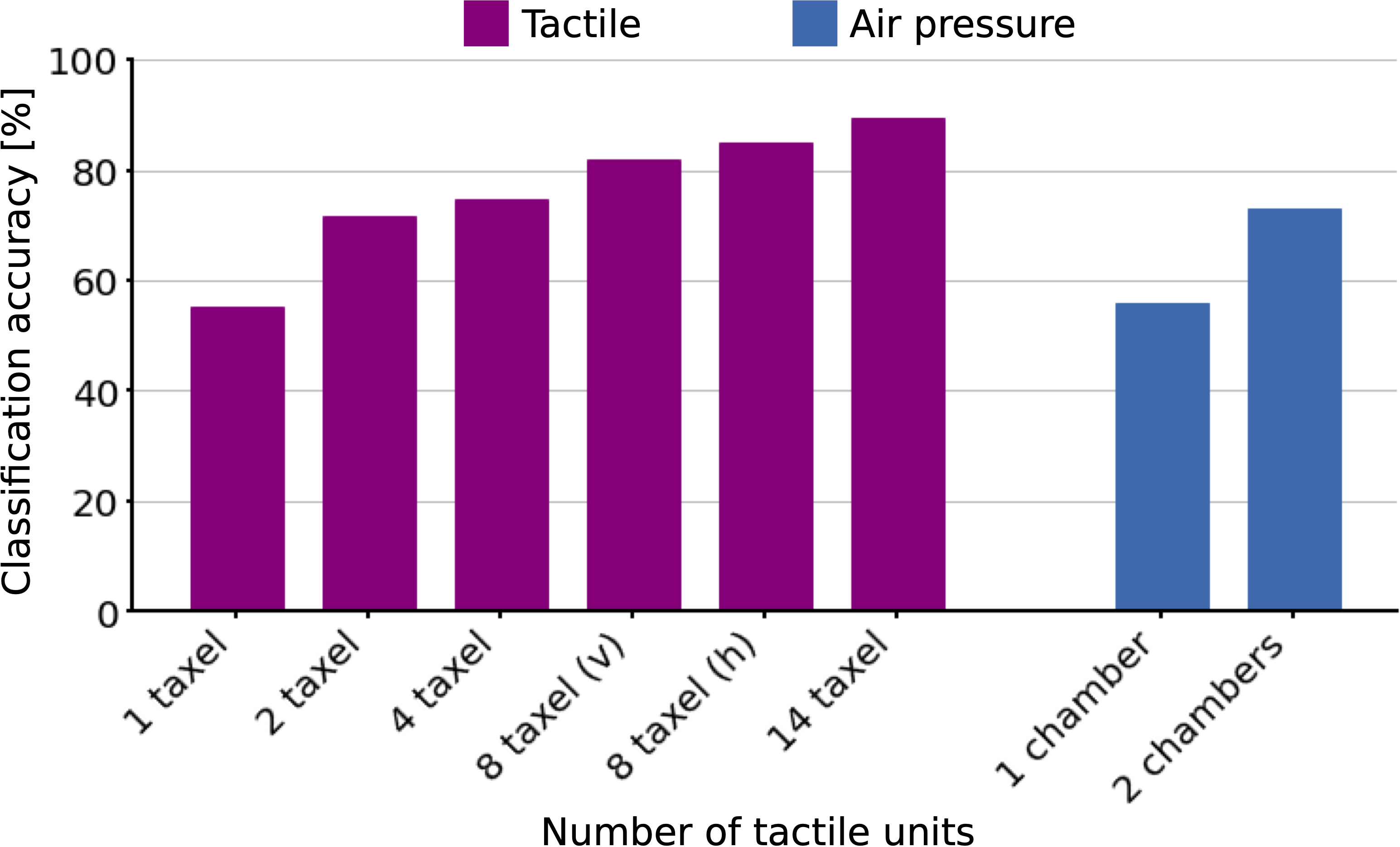}
  \caption{Object recognition performance improves for increasing
    spatial resolution of each finger's tactile sensing (violet) and
    air pressure sensing (blue).  Combining taxel values at the
    fingertip horizontally~(h) yields better results than
    vertically~(v).}
\label{fig:spatial-resolution}
\end{figure}

Object recognition performance for the different spatial resolutions of the two sensor modalities is shown in Figure~\ref{fig:spatial-resolution}.  Interestingly, the two sensor modalities achieve comparable performances for the same number of sensing units. For a single taxel, tactile sensing achieves 55.2\% and air pressure sensing 55.9\%. For two taxels, performance rises to 71.5\% and 72.9\%, respectively. This indicates the ability of pressure sensing to capture relevant contact information. However, the number of sensing units of these sensors is determined by the number of air chambers and their response is influenced by the inflation state of the soft actuator.

The results indicate that for both sensory modalities, more tactile units result in better classification performances. This, of course, is not surprising. However, it demonstrates our sensor's ability to provide rich sensory feedback despite its simple design. Furthermore, the results indicate that also different structure in combining the same number of taxels yields different results (compare vertically vs. horizontally combined fingertip taxels in Figure.~\ref{fig:spatial-resolution}) which is an interesting topic for future research. 

%\balance

%======================================================================
\section{Conclusion}
%======================================================================

We proposed a compliant sensor for soft robotics. Its design, based on thin and flexible materials, minimizes detrimental effects on morphology and compliance when attached to a soft actuator. The sensor features 14 taxels, realized by electrodes on a flexible printed circuit board. The sensor can be manufactured at low-cost in very easy steps within minutes.  We characterized the sensor and explained how choosing an appropriate pull-up resistance leads to adjustments in sensitivity and measuring range to match task requirements. The sensor was evaluated on the soft pneumatic RBO~Hand~2.  In this context, we showed that the sensor has only little effect on compliance. We also showed that our sensor provides useful tactile information by demonstrating the sensorized hand's capability to identify grasped objects. Finally, we compared classification performance based on tactile and air pressure information and showed that tactile sensing recognizes objects more successfully, thanks to its superior spatial resolution.

%======================================================================
\bibliography{library}{}
\bibliographystyle{ieeetr}
%======================================================================

\end{document}